\documentclass[letterpaper, 10 pt, conference]{ieeeconf}  

\IEEEoverridecommandlockouts                              
\usepackage{graphics} 
\usepackage{epsfig} 
\usepackage{mathptmx} 
\usepackage{times} 
\usepackage{amsmath} 
\usepackage{amssymb}  
\usepackage{algorithm,algorithmic}
\usepackage{scalerel}
\usepackage[dvipsnames]{xcolor}
\usepackage{cancel}
\usepackage{pifont}
\usepackage{tabularx,booktabs}
\usepackage{hyperref}
\usepackage{booktabs}
\usepackage{threeparttable}

\newcommand{\vn}[1]{{\bf #1}}
\newcommand{\norm}[1]{\ensuremath{\left \Vert #1 \right \Vert}}

\colorlet{Mycolor1}{green!10!orange!90!}

\newcommand{\NA}{---}
\newcolumntype{C}{>{\centering\arraybackslash}X} 

\title{\LARGE \bf
Look at my new blue force-sensing shoes!
}

\author{Yuanfeng Han$^{*}$,  Ruixin Li$^{*}$ and  Gregory S. Chirikjian$^{\dagger,*}$
\thanks{$^{*}$ Yuanfeng Han and Ruixin Li are with the Department of Mechanical Engineering, Johns Hopkins University, Baltimore, MD.
       {\tt\small yhan33@jhu.edu}}
\thanks{$^{\dagger,*}$ Gregory S. Chirikjian is with the Department of Mechanical Engineering,
National University of Singapore, Singapore and the Department of Mechanical Engineering, Johns Hopkins University, Baltimore, MD.
        {\tt\small mpegre@nus.edu.sg, gchirik1@jhu.edu}}%
}

\begin{document}

\maketitle
\thispagestyle{empty}
\pagestyle{empty}

\begin{abstract}
To function autonomously in the physical world, humanoid robots need high-fidelity sensing systems, especially for forces that cannot be easily modeled. Modeling forces in robot feet is particularly challenging due to static indeterminacy, thereby requiring direct sensing. Unfortunately, resolving forces in the feet of some smaller-sized humanoids is limited both by the quality of sensors and the current algorithms used to interpret the data. This paper presents light-weight, low-cost and open-source force-sensing shoes to improve force measurement for popular smaller-sized humanoid robots, and a method for calibrating the shoes. The shoes measure center of pressure (CoP) and normal ground reaction force (GRF). The calibration method enables each individual shoe to reach high measurement precision by applying known forces at different locations of the shoe and using a regularized least squares optimization to interpret sensor outputs. A $\text{NAO}^{\text{TM}}$ robot is used as our experimental platform. Experiments are conducted to compare the measurement performance between the shoes and the robot’s factory-installed force-sensing resistors (FSRs), and to evaluate the calibration method over these two sensing modules. Experimental results show that the shoes significantly improve CoP and GRF measurement precision compared to the robot’s built-in FSRs. Moreover, the developed calibration method improves the measurement performance for both our shoes and the built-in FSRs.

\end{abstract}

\section{Introduction} \label{intro}
Foot force sensors are widely used in humanoid robots for measuring GRFs and CoPs, which quantify the stability status of the system \cite{vukobratovic2004zero}. Many researchers and developers design planning and control algorithms for humanoid robots according to their foot sensory feedback in the studies of dynamic walking \cite{tsuichihara2011sliding}, self-balancing \cite{nakaura2002balance}, push-and-recovery \cite{ghassemi2014push}, etc. Additionally, many state estimation problems such as estimating center of mass (CoM) location \cite{hawley2016external} and external forces \cite{piperakis2018nonlinear} in the dynamic motion of the humanoid robot require the integration of foot force data together with other in-body sensory data like an encoder or an IMU. Recent studies also show that high-precision foot force sensors can be used to identify the inertial properties of heavy objects for humanoid robot manipulation tasks \cite{han2020can}. 
\begin{figure}[t!]
\centering
\includegraphics[width=0.7\linewidth]{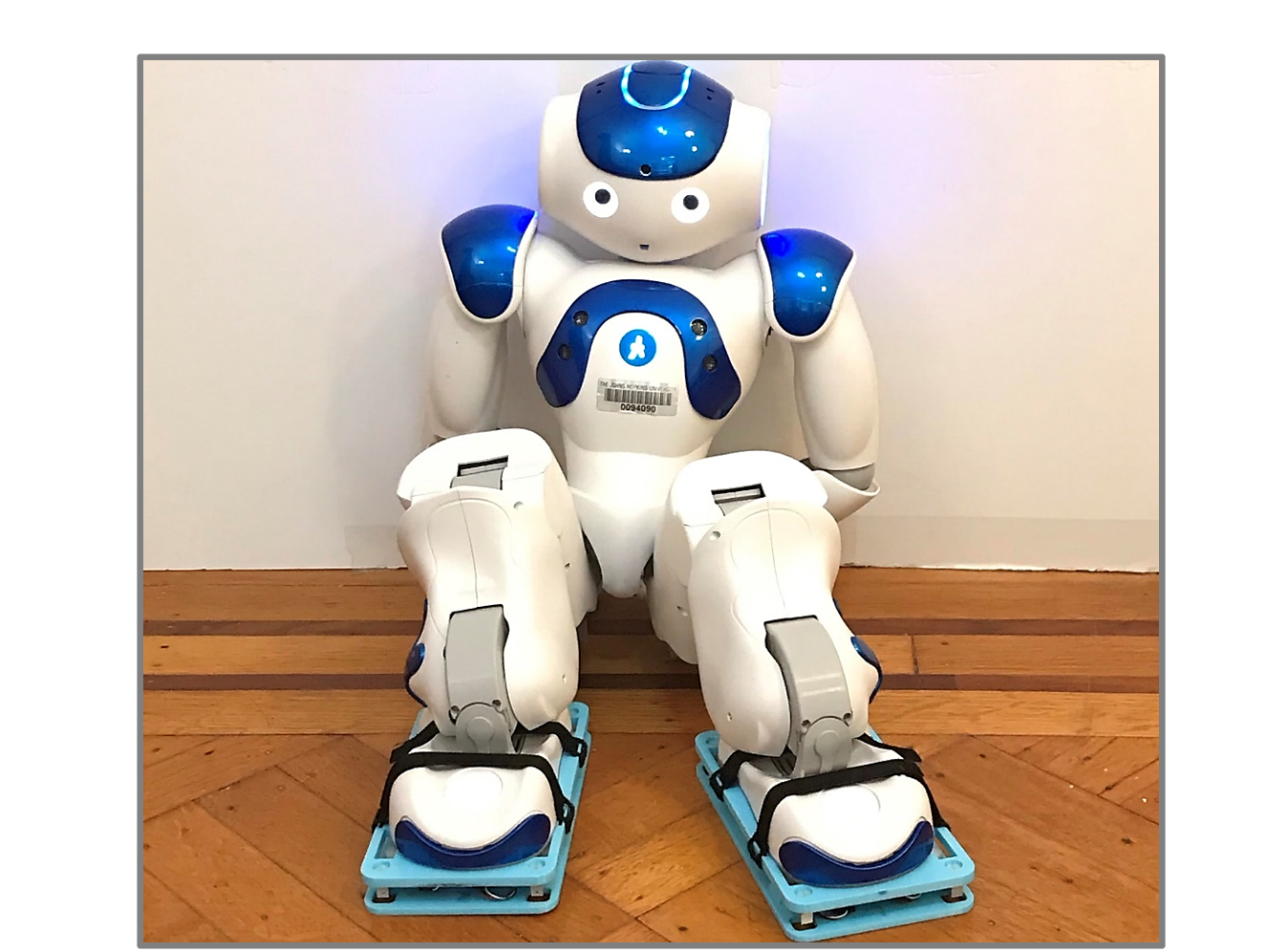}
\caption{A Nao robot wears a pair of new force-sensing shoes.}
\end{figure}

Two major types of force sensors are commonly used in measuring CoPs and GRFs for humanoid robots. Commercial six-axis force/torque (F/T) sensors have been applied for many larger-sized humanoid platforms \cite{takenaka2006control, koch2014optimization}. This type of sensor utilizes a strain gauge mechanism, which outputs a linear response to the external force and possesses high measurement precision. However, six-axis F/T sensors are generally expensive and heavy -- the added weight increases the moment of inertia of the foot and decreases the dynamic postural stability \cite{fujimoto1998attitude}. Therefore, six-axis F/T sensors have limited applications in the context of smaller-sized humanoid platforms. Another common type of foot sensors are thin-film FSRs: these are usually installed on relatively smaller humanoid platforms, such as the $\text{NAO}^{\text{TM}}$ (subsequently this is referred to as Nao) \cite{gouaillier2009mechatronic} and the H7 robot \cite{takahashi2005high}. Despite their light-weight nature, FSRs generally have lower measurement precision and repeatability compared to strain-gauge type sensors. Therefore, without effective calibration methods, it is difficult to use such sensors for estimating accurate CoP location in the feet of humanoid robots.

The performance of force sensors can be affected by various factors. For example, changes in the ambient temperature affects the physical properties of the strain gauge \cite{chavez2019model} and the resistance coefficient of the FSRs \cite{paredes2017framework}, causing the sensor to drift. Moreover, slight variations in the mechanical structure during usage, such as changes in screw tension and deformations of the mounting material, may also produce drift or alter sensor output. Therefore, an effective calibration method is critical for improving measurement precision of foot force-sensing modules.

There exist several studies focusing on developing foot force-sensing modules for CoP and GRF measurements for smaller-sized humanoid robots. Shayan et al. presented a modular shoe design for the Nao using barometric sensors \cite{shayan2019design}. Almeida et al. proposed a force-sensing shoe design using two separate layers for the Nao \cite{almeida2018novel}. Kwon et al \cite{kwon2011fabrication} developed a pressure-sensing foot using custom-fabricated polymer-based film for a KIBO robot. Suwanratchatamanee et al \cite{suwanratchatamanee2009haptic} developed haptic-sensing feet with three sensing layers for a KHR-2HV robot. The previously developed foot-sensing modules can theoretically improve measurement precision with their high-quality sensors. However, a detailed calibration technique is necessary to further optimize their sensing performance and evaluation experiments are needed to quantify the sensing precision of these modules.

This paper presents a pair of low-cost, light-weight and open-source force-sensing shoes for smaller-sized humanoid robots (Fig. 1). The new shoes have many advantages over the above-mentioned existing foot-sensing modules (see Table I). A calibration method is proposed, which utilizes regularized least-squares to minimize the error between the measured and the reference CoPs and GRFs using our 3-D printed calibration apparatus. The calibration enables the shoes to reach high measurement precision. For each calibrated shoe, the CoP mean absolute error is less than 2 mm and the GRF mean absolute error is less than 0.025 N. Three validation experiments are implemented on both the Nao's factory-installed FSRs in its feet and our new shoes. The results show that the shoes possess a significantly higher measurement precision compared to the built-in FSRs of the Nao; The developed foot-sensor calibration method improves the CoP and GRF measurement precision for both the built-in FSRs and the force-sensing shoes.

\section{Nao's foot and force-sensing shoe}
The CoP and GRF sensing performances are compared between the Nao's factory-installed FSRs in its feet and our new force-sensing shoes. 

\subsection{Nao's foot} \label{foot}
Each foot of the Nao contains four FSRs (Fig. 2a), of which the working range are 0 \-- 25 N. These FSRs form an approximate 100 $\times$ 53 mm rectangular sensing area (Fig. 2b) (referred to as foot sensing area), which defines the stable region of the CoP inside each foot. The bottom of the foot is covered by a rigid plate, which connects to the top foot via screws (Fig. 2a). The CoP and GRF measurements of the foot rely on the deformation of the bottom plate to deliver force to each FSR. However, the measurement is unreliable since the bottom foot plate barely deforms and it is unlikely that the four FSRs will activate at the same time.
\begin{table}[t!]
\caption{} 
\centering 
\begin{tabular}{c c c c c c c} 
\hline\hline 
\text{Foot} & \text{Light} & \text{Low} & \text{Off-the-shelf}  & \text{3D} & wireless & \text{open}\\
\text{module} & \text{weight} & \text{cost} & \text{component}  & \text{printable} &  & \text{source}\\
\hline 
Nao Pressure\cite{shayan2019design}    & \ding{51} & \ding{51} & \ding{55} & \ding{51} & \ding{51} & \ding{55} \\
Nao ITShoe\cite{almeida2018novel}      & \ding{51} & \ding{51} & \ding{51} & \ding{55} & \ding{51} & \ding{55} \\
H7 Matrix FSR\cite{takahashi2005high}   & \ding{51} & \NA       & \ding{55} & \ding{55} & \ding{55} & \ding{55} \\
KIBO Foot\cite{kwon2011fabrication}       & \ding{51} & \NA & \ding{55} & \ding{55} & \ding{55} & \ding{55} \\ 
KHR-2HV Foot\cite{suwanratchatamanee2009haptic}    &\ding{51}  & \NA       & \ding{55} & \ding{55} & \ding{55} & \ding{55} \\ 
Our Shoe        & \ding{51} & \ding{51} & \ding{51} & \ding{51} & \ding{51} & \ding{51} \\
\hline 
\end{tabular}
\begin{tablenotes}
\item \NA \quad property is not is introduced in the literature
\end{tablenotes}
\label{table:smallsensor} 
\end{table}
\begin{figure}[t!]
\centering
\includegraphics[width=1.0\linewidth]{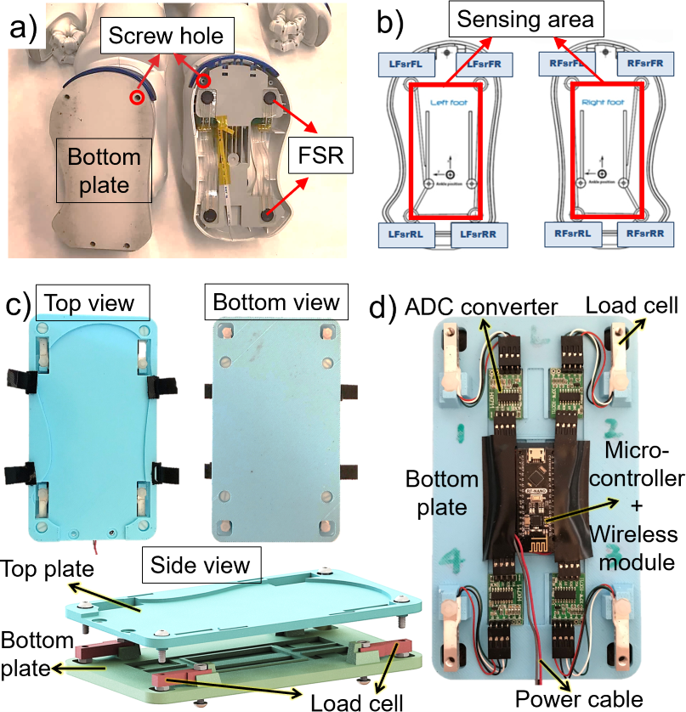}
\caption{a) Nao's foot design. b) Sensing area of the foot. c) Shoe schematics. d) Shoe's electronic components.}
\end{figure}
\subsection{Force-sensing shoe}
A pair of force-sensing shoes are developed for smaller-sized humanoid robots. Single-axis load cells are chosen to construct the shoes due to their high-accuracy, linear-response and low-drift. In addition, small-size, light-weight and low-cost single-axis load cells are readily available commercially. Other options such as soft-matter based sensors and optical sensors are also considered in the sensor selection process. However, soft-matter based sensors are often susceptible to hysteresis, which limits their accuracy when measuring rapidly or cyclically changing forces \cite{park2012influence}; Relatively accurate optical-based force sensors can be costly and often require complex setups \cite{almassri2015pressure}. Here, the working range of our selected load cells are 0 \-- 50 N. The dimensions of the assembled shoes are $170 \textrm{mm} \times 95 \textrm{mm} \times 21 \textrm{mm}$. Each shoe weighs 135g -- only $2.4\%$ of the mass of the Nao. The overall component cost of each shoe is less than 30 US dollars. The shoe is fully open source and the details are in \textit{https://github.com/yhan33/Force-sensing-shoes-for-Nao}. Our force-sensing shoes possess many advantages over the existing foot force-sensing modules introduced in section \ref{intro}, which are shown in Table I.
\subsubsection{Mechanical Design}
Each shoe works similarly like a force plate. The shoes are comprised of two 3-D printed plates. The upper plate has an indentation which tightly fits the Nao's foot (Fig. 2c, top view). The electronic components are mounted to the bottom plate and sit under the upper plate (Fig. 2c, d). For each load cell, its one side is cantilevered from the pillar on the bottom plate and the tip of the other side connects to the top plate via a screw (Fig. 2c, side view). The top plate is designed much thinner and more flexible compared to the thick and rigid bottom plate. This design ensures that all four load cells engage simultaneously when the robot puts weight on the shoes. Since the whole bottom plate contacts with the ground, the shoe can also be used for measuring forces on uneven terrain. In contrast, in designs in which four rigid load cells are directly on the flat floor, it is very likely that one of them may lose contact due to manufacturing or assembly misalignment. In addition, such design can hardly adapt to uneven terrain since sensors may lose contact if the middle area of the shoe is lifted up by obstacles. The shoes are designed with straps to facilitate easy removal. In practice, they can be screwed or glued on the foot of the robot for common usage.

\subsubsection{Data transmission}
All electronic components in the force-sensing shoes are chosen off-the-shelf (Fig. 2d). The shoes can be powered by external batteries. Raw data from the load cells first passes through the corresponding Analog-to-Digital Converters, which is then transmitted wirelessly by a micro-controller. A sister module receives the data and transfers it to a PC via USB serial communication. The communication frequency between the shoes and the PC can reach up to 80 Hz. 

\section{Foot and shoe calibration}
The calibration algorithm developed here is tested both on the Nao's built-in FSRs and on our force-sensing shoes.
\subsection{Measurement principle}\label{measure}
Both the Nao's foot and the shoe incorporate four individual sensors' data to obtain the CoP and GRF measurement. We denote each sensor's force output as $F_{i}$ and its location in $x$-$y$ plane as seen from the local frame attached to the foot or the shoe as $\mathbf{p}_{i}\in \mathbb{R}^{2}$ ($i= 1, \, \ldots, \, 4$). The measured GRF value and CoP location for each foot or shoe are given by $\sum_{i} F_{i}$ and $\sum_{i} F_{i} \, \, \mathbf{p}_{i}  \, / \, \sum_{i} F_{i}$, respectively.  The CoP and GRF measurements for double support can be obtained by simply summing over all eight sensors.

\subsection{FSR and load cell force output} \label{single_cal}
\subsubsection{Nao's foot FSR}
The factory installed FSRs are calibrated before assembly such that their raw sensor outputs directly indicate the forces applied to the sensors.

\subsubsection{Single load cell}
Each load cell is calibrated to convert raw sensor output to physical force. Here, we denote the loaded raw sensor reading as $S$, the average unloaded reading as $a$, the scaling factor as $b$, and the force measurement as $F$. The mapping between raw sensor reading and force output can be described using an affine transformation:
\begin{equation}
F = \frac{(S-a)}{b} = \frac{S}{b} - \frac{a}{b} = cS+d, \label{affine}
\end{equation}
in which $c$ and $d$ are the modified scaling factor and offset. The offset $a$ is obtained by recording the average no-load reading of the sensor. The scaling factor $b$ is computed by first placing a known weight on the sensor and then dividing the change in each sensor's output by the weight's gravitational force.

\subsection{Foot and Shoe calibration method} \label{shoe_cal}
The Nao's foot does not deliver accurate CoP and GRF measurements (see \ref{foot}). Similarly, the shoe's measurement precision is likely to drop due to mechanical misalignment and internal deformations or stresses after assembly. Therefore, it is necessary to calibrate the sensor parameters for both feet and shoes to enable optimal measurement precision.

The calibration apparatus is shown in Fig. 3, top. A 3D-printed sole plate is attached to the foot or the shoe. The sole plate has a 3 by 6 array of cylindrical protrusions designed to support a variable weight. The exact location of each protrusion relative to the foot's or shoe's body frame is pre-recorded, and are used to obtain the reference CoP location. Since the whole calibration apparatus is static
during calibration process, the reference CoP can be obtained by projecting the CoM of the calibration apparatus onto the $x$-$y$ flush with foot or the shoe. The reference GRF is simply the gravitational force of the calibration apparatus. Denoting the reference CoP location as $[C_{x}, \, C_{y}]^T \in \mathbb{R}^2$ and the reference GRF as $N$, they are given by:
\begin{align}
\begin{bmatrix}
\text{C}_{x}\\
\text{C}_{y}\\
\end{bmatrix}=\frac{(G_{\text{weight}} + G_{\text{cap}})\begin{bmatrix}
p_{x}\\
p_{y}
\end{bmatrix} + G_{\text{sole}}\begin{bmatrix}
p_{x}^{s}\\
p_{y}^{s}
\end{bmatrix}}{G_{\text{weight}} + G_{\text{cap}} + G_{\text{sole}}},
\end{align}
\begin{align}
N = G_{\text{weight}} + G_{\text{cap}} + G_{\text{sole}},
\end{align}
where $G_{\text{weight}}, \, G_{\text{cap}}, \, G_{\text{sole}}$ are the weight of the calibration weight, the supporting cap and the sole plate\footnote{For the Nao's foot, experiments show that the weight of its bottom plate does not impact the sensor outputs when the foot is upside down.} (Fig. 3b, top, right) and $[p_{x}, \, p_{y}]^T$ is the reference CoP location of the weight and supporting cap together, which is the center of the protrusion used in the measurement. Here, $[p_{x}^{s}, \, p_{y}^{s}]^T$ are the CoM of the sole plate, which is obtained from its CAD model. The calibration algorithm aims to update sensor parameters to enable optimal measurement precision. Here, the raw sensor output of each FSR and load cell is denoted as $S$, and the corresponding updated force output $F$ is modeled using an 1-D affine transformation such that $F = cS+d$. Denoting the measured CoP location in the $x$, $y$ directions as $[c_{x}$, $c_{y}]^T$ and the measured GRF as $n$, the measured values are given by:
\begin{align}
\begin{bmatrix}
\text{c}_{x}\\
\text{c}_{y}\\
\end{bmatrix}=
\frac{\sum_{k = 1}^{4}(F_{k}\begin{bmatrix}
t_{x}^{k}\\t_{y}^{k}
\end{bmatrix})}{\sum_{n = 1}^{4}F_{k}}
=\frac{\sum_{k = 1}^{4}(c_{k}S+d_{k})\begin{bmatrix}
t_{x}^{k}\\t_{y}^{k}
\end{bmatrix}}{\sum_{k = 1}^{4}(c_{k}S+d_{k})},
\end{align}
\begin{align}
n = \sum_{k=1}^{4}F_{k} = \sum_{k=1}^{4}(c_{k}S+d_{k}),
\end{align}\label{singlemeasure}
(see \ref{measure}), in which $k$ represents the designated sensor and $[t_{x}, t_{y}]^T$ indicate the location of the sensor's measuring point in the $x$-$y$ coordinate of the frame of the foot or the shoe.

\begin{figure}[t!]
\centering
\includegraphics[width=1.0\linewidth]{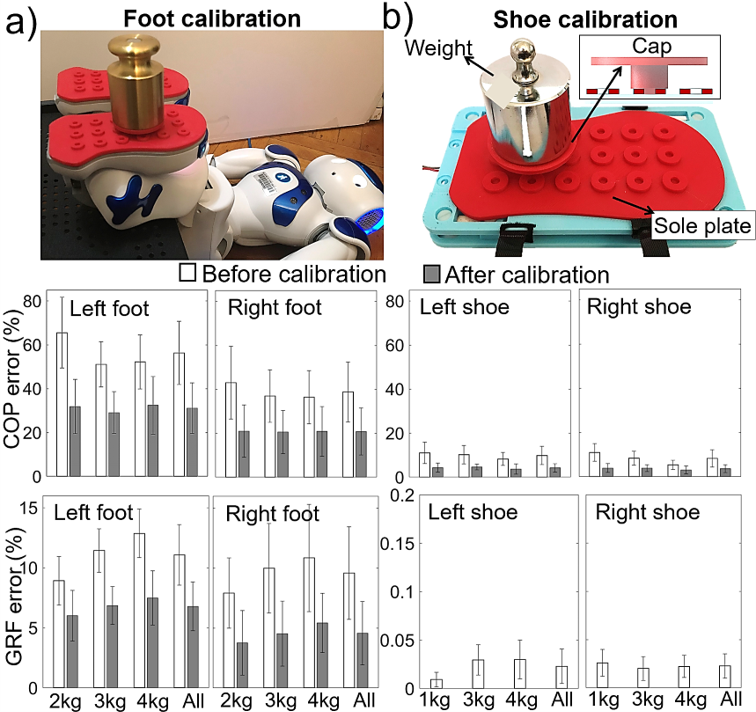}
\caption{a) Foot calibration. b) Shoe calibration. Middle row shows the CoP relative measurement error of the left and right foot/shoe. Bottom row shows the GRF relative measurement error of the left and right foot/shoe. The empty and filled bars show the error before and after calibrations. The GRF measurement is not calibrated for the Shoe. Note that scales are different so as to accurately display data.}
\end{figure} 

\begin{table}[htbp]
\caption{Mean absolute measurement error} 
\centering 
\begin{tabularx}{0.45\textwidth} { 
    |>{\centering\arraybackslash}X | >{\centering\arraybackslash}X 
    >{\centering\arraybackslash}X 
    >{\centering\arraybackslash}X 
    >{\centering\arraybackslash}X| }
 \hline
         & CoP MAE (mm) & CoP MAE (mm) & GRF MAE (N)  & GRF MAE (N)\\
 \hline
Calibration  & \text{NO}  & \text{YES} & \text{NO}  & \text{YES} \\
 \hline
 Left Foot  & 23.13 & 12.81 & 10.85 & 6.65\\
\hline
 Right Foot  & 15.95 & 8.56 & 9.39  & 4.48\\
 \hline
Left Shoe  & 4.07 & 1.76 & 0.023  & \text{N/A}\\
\hline
 Right Shoe  & 3.41 & 1.51 & 0.022  &\text{N/A}\\
\hline
\end{tabularx}
\end{table}

A least-squares regression is used to minimize the difference between the reference CoP and GRF and their measured values. To prevent overfitting, an extra regulation term is applied to penalize the difference between the updated force output and the initial force output. Specifically, the updated affine model parameters of the load cell are bounded around the initial calibrated parameters (see \ref{single_cal}) and the updated FSR force output is bounded around the raw force output.

Since the FSRs exhibit poor CoP and GRF measurement performance (Fig. 3, left, empty bars), the optimization is designed to minimize both CoP and GRF error over the space of 1-D affine transformations. The regularized least square problem is formulated as:
 \begin{align}
     \underset{\mathbf{\zeta}}{\text{argmin}} \ \vn{J}  = \left(\norm{\vn{C}_x - \vn{c}_x}^2 + \norm{\vn{C}_y - \vn{c}_y}^2\right)_{\vn{w}_c} \label{cost} \\ 
      + \norm{ \vn{N} - \vn{n}}^2_{\vn{w}_n} + \norm{\zeta - \zeta_0}_{\vn{w}_\zeta}^2 \nonumber,
\end{align}
in which optimization variables $\zeta = [c_{1},\ldots,c_{4}, \ d_{1},\ldots,d_{4}]$. $\mathbf{\zeta_{0}}$ represent the FSRs' initial force outputs. $\mathbf{C}_{x},\mathbf{C}_{y}, \mathbf{c}_{x}, \mathbf{c}_{y}, \mathbf{N}, \mathbf{n}$ are vectors corresponding to the measured and reference CoP and GRF across all different calibrating locations and weights. Here, $\mathbf{w}_{c},\mathbf{w}_{n},\mathbf{w}_{\xi}$ are the weights associated with each set of terms. 

Compared with the FSRs, the shoes' GRF measurement is very precise (Fig. 3b, bottom, empty bars). Therefore, the shoe's GRF measurement can be obtained directly from the initial calibration parameters of the load cells $\zeta_0$ (see \ref{single_cal}) and the updated parameters $\zeta$ are used only for CoP measurement. Therefore, the weight term $\vn{w}_n$ for minimizing GRF error is set to $\vn{0}$ for the shoe's calibration.

\subsection{Performance evaluation}\label{error}
To ensure our calibration covers a broad sensing range, 1 kg, 2 kg and 4 kg weights ($\approx 18\%\textrm{, }36\%$ and $74\%$ of the robot's mass) are placed on each protrusion in the $3 \times 6$ array (Fig. 3b) for the shoe's calibration; 2 kg, 3 kg (55\% of the robot's mass) and 4 kg weights are placed on the middle three rows of the protrusions for the foot's calibration\footnote{Some FSRs do not respond to 1 kg weight in our experiment and only three middle rows are within the sensing area confined by the FSRs.}. The manual calibrations of the foot and the shoe comprise 36 and 54 different combinations of weights and positions.

Two relative error metrics, $e_{C}$ and $e_{G}$, are designed to quantify the precision of the CoP and GRF measurements:
\begin{gather}
e_{C} = \frac{\pi(||\mathbf{C}_{x} - \mathbf{c}_{x}||^{2}+||\mathbf{C}_{y} - \mathbf{c}_{y}||^{2})}{A} , \label{eCoP} \\
e_{G} = \frac{||\vn{N}-\vn{n}||}{\sum_{i=1}^{4}G_{fsr}}, \label{eGRF}
\end{gather}
where $A$ is the area of the support polygon and $G_{fsr}$ is the sensing range of each FSR.  Here, $e_{C}$ expresses the distance between the reference and measured CoP, relative to the radius of a circle whose area is equivalent to that of the foot sensing area, and $e_{G}$ represents the absolute GRF error over the total sensing range of one foot. For comparison purpose, the denominators of the error metrics of the shoe are the same as those of the foot.

The error metrics corresponding to each chosen weight averaged over different locations in the calibration array and the error metrics overall the weights and locations of the left and right feet and the left and right shoes are shown in Fig. 3. The overall errors for both CoP and GRF (only for foot) are reduced after calibrations (Fig. 3, filled bar) and remain relatively stable across different weights, suggesting that the optimized parameters produce accurate measurements over a wide range of forces. In addition, the force-sensing shoes show significant higher precision compared to the feet of the Nao both before and after calibrations.

The mean absolute error (MAE) of the feet and the shoes are also shown in TABLE II. The CoP and GRF MAE are defined by the averaged Euclidean distance between the measured and the reference CoPs and the averaged absolute difference between the measured and reference GRFs. For each shoe, the MAE of CoP (after performing calibration) is less than 2 mm and the MAE of GRF is less than 0.025N.

\begin{figure}[t!]
\centering
\includegraphics[width=0.99\linewidth]{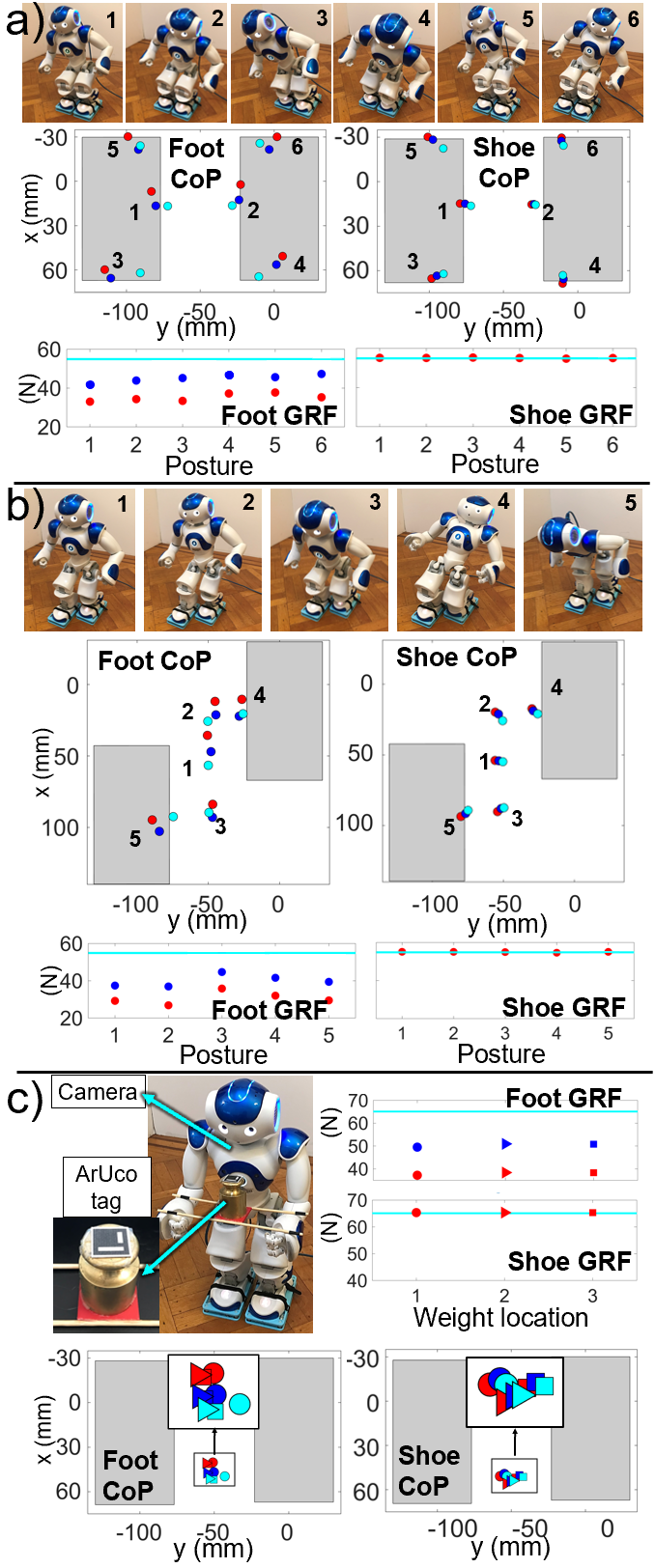}
\caption{Evaluation experiments. a, b top show designed configurations and c top-left shows one demo of the weight experiment. a, b middle and c, bottom show CoP measurement of the feet (left) and shoes (right). a, b bottom and c top-right show GRF measurements. Teal, red and blue show the reference value, measurement value without calibration and with calibration. Circle, square and triangle in c represent different weight placements.}
\end{figure}

\section{Experiments} \label{exp}
A Nao robot is used as our testing platform. Experiments are designed to evaluate the precision of CoP and GRF measurements using error metrics defined in \ref{error} by the Nao's built-in FSRs and the shoes using two force readings: 1. the raw sensor force readings without calibration (see \ref{single_cal}); 2. the updated force readings after calibration (see \ref{shoe_cal}). The experiments are performed using several pregenerated double support configurations of the robot and the measured results are eventually compared with the reference results (see section \ref{Res}).  The experiments are explained as follows:\newline
\textbf{1. Edge and center test. }
The CoP and GRF are measured using six different generated postures in a fixed double support. The CoPs of these postures are distributed in different locations in the support polygon (Fig. 4a). Four of the postures are near the edge of the support polygon (Fig. 4a, 3 - 6) and two are close to the middle (Fig. 4a, 1, 2). In the experiment, robot remains static in a posture for 8 seconds while sensing data is recorded and then averaged and evaluated. Fig. 4a shows one set of experimental data of CoP (middle) and GRF (bottom) measurement for the feet (left) and the shoes (right). 
\newline
\textbf{2. Changed double support test. }
The CoP and GRF measurements are tested in five postures (Fig. 4b, 1-5) in a very different double support compared to the one designed in the previous experiment (Fig. 4a): the robot moves its right foot 70 mm to the front (close to the robot's maximum step length). The testing details are the same as the first experiment and one set of result is shown in Fig. 4b.
\newline
\textbf{3. External weight test. }
To evaluate foot sensing performance reacting to external forces, an external 1 kg weight with an attached ArUco tag \cite{garrido2014automatic} is placed at three random locations on the robot's arms within the detection range of the robot's camera (Fig. 4c, top-left). The testing details are the same as the previous experiments and one set of result is shown in Fig. 4c.

For comparison purpose, both the feet and the shoes are tested using the same double support configurations. The grey areas in Fig. 4 are the sensing areas of the feet (see Fig. 2b). Each of the three experiments described above is performed on three separate days for the Nao's feet, and on three separate days for the shoes (the shoes are not detached during the shoes' experiments). The calibration is only performed for the first day and the updated parameters are used for the experiments on the following two days. The averaged results of each individual experiment are used for measurement performance evaluation.

\begin{figure}[t!]
\centering
\includegraphics[width=1.0\linewidth]{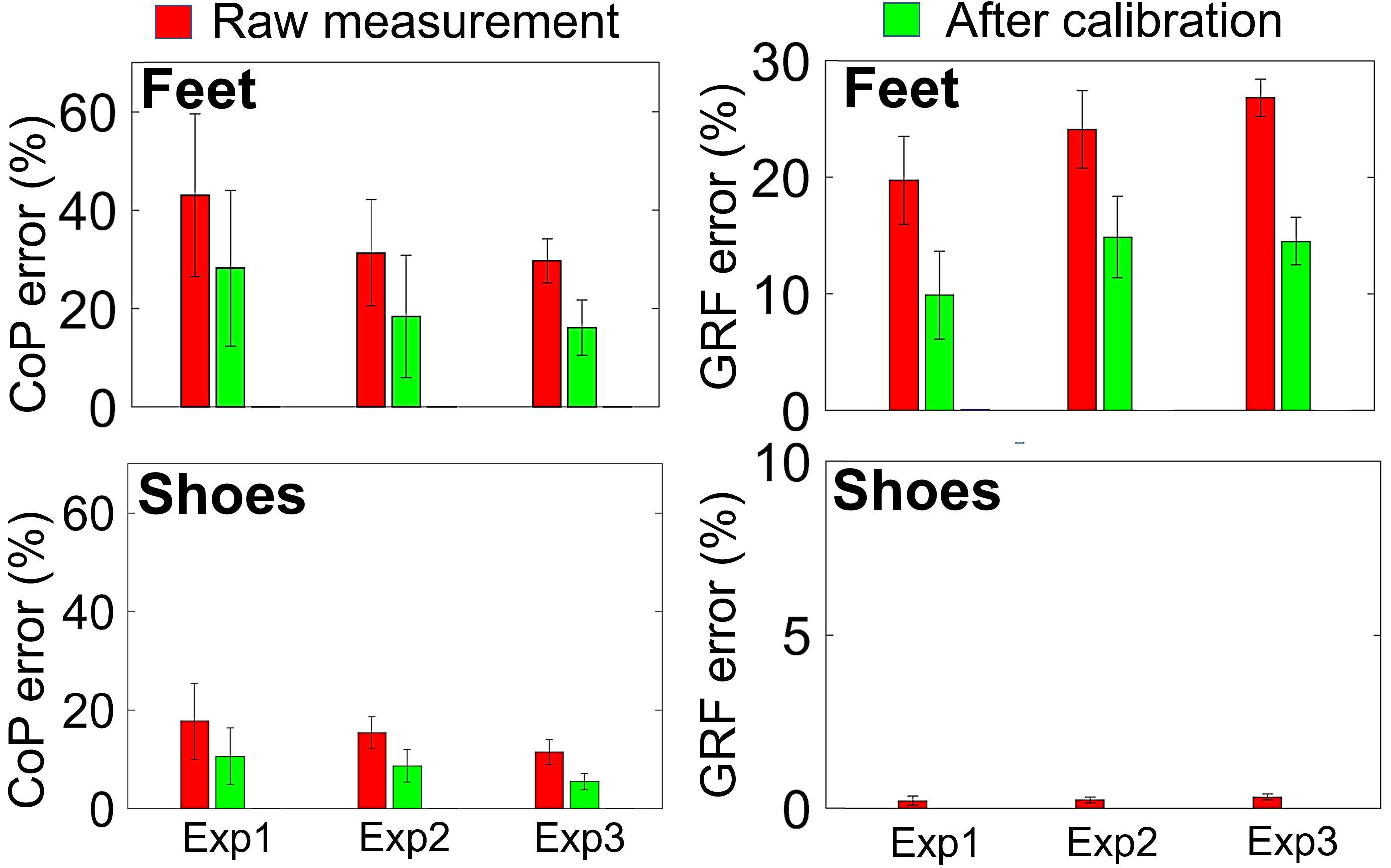}
\caption{CoP and GRF relative errors of three evaluation experiments. The CoP (left) and GRF (right) of the Nao's feet (top) and the shoes (bottom). Red and green show results with and without calibration. Scales are different so as to accurately display data.}
\end{figure} 

\section{Results} \label{Res}
To evaluate the measurement performance, the reference CoP and GRF corresponding to each measured result is required. Since the inertial properties of the Nao is fully provided by its company, which showed descent accuracy in our previous work \cite{han2020can}, here we use modeled CoP and GRF as references: when the robot stands statically, its reference CoP can be obtained using the projection of its CoM on the ground. In practice, the CoM of the robot is obtained by computing the CoM location of each joint and link relative to the robot's body frame using encoder data and then implement a weighted average over the masses and positions of the links and joints. For Experimental 3, the CoM location of the weight relative to the tag frame is premeasured and the Nao uses its bottom camera to localize the pose of the tag to obtain the CoM of the weight in the robot's body frame, then the CoM location of the whole system can be computed accordingly. The reference GRF is simply the gravitational force of the whole system.

The overall experimental results are shown in Fig. 5. The red bars denote the CoP and GRF measurement errors (defined in \ref{error}) when calibrations are not performed and the green bars when calibrations are performed. The three columns show the results of the three evaluation experiments in \ref{exp}. The results show that our force-sensing shoes possess significant higher measurement precision (Fig. 5, bottom) compared to the robot's FSRs (Fig. 5, top) across all experiments. This is likely due to the design of the shoes, which ensures the four load cells activate simultaneously, in contrast to the rigid plates fixed to the Nao's feet, which hinders simultaneous activation of the FSRs. In addition, the calibrations (green) improve the CoP measurement precision for both the Nao's FSRs and the shoes, and the GRF measurement for the FSRs. The results also show that the highest CoP sensing precision is achieved when the Nao wears the calibrated force-sensing shoes (Fig. 5 bottom-left, green).

\section{Conclusion}
{
This paper presents a pair of low-cost, light-weight and open-source force-sensing shoes for CoP and GRF measurements for smaller-sized humanoid robots. A calibration procedure is introduced which utilizes regularized least-squares to obtain optimal sensor parameters. Evaluation results show that our new force-sensing shoes exhibit significantly higher measurement precision compared to the Nao's factory-installed FSRs in its feet. In addition, the developed calibration method is capable of improving CoP and GRF measurement precision for both the Nao's FSRs and the force-sensing shoes. The best performance occurs when the robot wears the calibrated shoes. In principle, both the shoes and the calibration method can be applied individually to many existing humanoid robots and force-sensing modules. 
}


\bibliographystyle{IEEEtran}
\bibliography{ICRA.bib}
\end{document}